\documentclass[11pt]{article}

\usepackage[final]{acl}

\usepackage{times}
\usepackage{latexsym}
\usepackage[normalem]{ulem}
\useunder{\uline}{\ul}{}
\usepackage[inline]{enumitem}
\usepackage{enumitem}
\usepackage{mathptmx}
\usepackage{amsmath}
\usepackage{amssymb}
\usepackage[T1]{fontenc}

\usepackage[utf8]{inputenc}

\usepackage{microtype}

\usepackage{inconsolata}

\usepackage{graphicx}
\usepackage{amsmath}

\usepackage[utf8]{inputenc}
\usepackage{booktabs} %
\usepackage{multirow} %
\usepackage{geometry} %
\title{FlowRAG: Synergizing Explicit Reasoning via Frequency-Aware Multi-Granularity Graph Flow}

\author{
  \textbf{Bihao Zhan\textsuperscript{1}\footnotemark[2]},
  \textbf{Zongsheng Cao\textsuperscript{2}\footnotemark[2]},
  \textbf{Jie Zhou\textsuperscript{1}\footnotemark[1]},
  \textbf{Bo Zhang\textsuperscript{2}\footnotemark[1]},
  \textbf{Liang He\textsuperscript{1,2}}
\\
  \textsuperscript{1}East China Normal University, \textsuperscript{2}Shanghai Artificial Intelligence Laboratory
}

\begin{document}
\maketitle

\footnotetext[1]{\textsuperscript{$\dagger$}Equal contribution. \quad}
\footnotetext[2]{Corresponding authors, jzhou@cs.ecnu.edu.cn, zhangbo@pjlab.org.cn}

\begin{abstract} 
Graph-based retrieval-augmented generation (GraphRAG) is effective for knowledge-intensive and multi-hop query tasks; however, many existing methods primarily seed entity-based graphs and rely on implicit semantic relevance propagation. This often (i) under-retrieves when user queries are abstract and semantically sparse at the entity level, and (ii) suffers from brittle multi-hop reasoning, where noisy activations can derail entity-to-entity transitions and corrupt the inferred relation chain, yielding unreliable conclusions. To this end, we propose \texttt{FlowRAG}, a semantic-aware retrieval framework that improves both semantic recall and explicit reasoning. Specifically, \texttt{FlowRAG} constructs a quad-level heterogeneous graph over passages, summaries, sentences, and entities, where summary nodes serve as a coarse semantic hub. At retrieval time, a dual-granularity activation module combines summary--query alignment with sentence-level matching to activate relevant entities under paraphrase and abstraction robustly. We then introduce a frequency-aware weighted flow module that routes relevance through entity--passage links weighted by within-passage term frequency, pruning noisy connections and extracting high-confidence reasoning paths as an explicit logic skeleton for generation.  Extensive experiments show that \texttt{FlowRAG} obtains state-of-the-art performance on complex reasoning benchmarks.

\end{abstract}

\section{Introduction}
Recently, Retrieval-Augmented Generation (RAG)~\citep{lewis2020retrieval,cao2026vig} has emerged as a crucial paradigm for enhancing Large Language Models (LLMs) by grounding their responses in external knowledge bases, thereby mitigating hallucinations and improving factual accuracy. While traditional RAG systems excel at simple queries, they often falter when confronted with complex, multi-hop reasoning tasks over large-scale, unstructured corpora where relevant information is fragmented and distributed~\citep{edge2025localglobalgraphrag}. 


\begin{figure}
\centering
\includegraphics[width=0.48\textwidth]{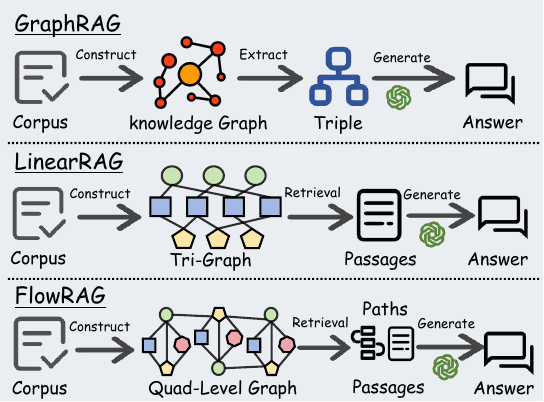} 
\caption{Comparison with other retrieval Methods. (i) GraphRAG relies on complex triple extraction. (ii) LinearRAG employs static ranking to return disjointed passages. (iii) \texttt{FlowRAG} (Ours) utilizes weighted flow to extract explicit reasoning paths rather than unstructured text segments.}
\label{fig:intro}
\vspace{-3mm}
\end{figure}

To address the limitations of flat retrieval in capturing structural dependencies, a diverse landscape of GraphRAG methods has emerged~\citep{han2025rag}. As shown in Figure \ref{fig:intro}, early approaches constructed explicit knowledge graphs via Open Information Extraction (OpenIE) to enable multi-hop traversal~\cite{martinez2018openie}. Evolving from this, systems like G-Retriever~\cite{he2024gretriever} leverage graph neural networks to target relevant subgraphs, while RAPTOR~\cite{sarthi2024raptor} introduces recursive clustering to build hierarchical tree structures for high-level context. Recent innovations, such as HippoRAG~\cite{jimenez2024hipporag} and LightRAG~\cite{guo2024lightrag}, further utilize associative indexing or dual-level retrieval strategies to uncover hidden connections. However, these methods often incur high computational costs due to complex extraction pipelines. To mitigate these engineering bottlenecks, LinearRAG~\cite{zhuang2025linearraglineargraphretrieval} was recently proposed, introducing a streamlined, relation-free hierarchical graph to significantly improve retrieval efficiency.

Despite these advancements, existing graph-based methods still face two fundamental challenges when handling complex reasoning tasks. The first challenge is under-retrieval due to entity sparsity. When user queries are abstract or semantically sparse (\textit{i.e.}, lacking explicit entity mentions that match the graph nodes), models often fail to activate the correct entry points, leading to a "granularity mismatch" between high-level query themes and fine-grained graph evidence. The second challenge is error propagation caused by noise. During multi-hop reasoning, noisy activations in the query or irrelevant connections within the graph can derail entity-to-entity transitions. Current methods often lack robust filtering mechanisms, causing the retrieval process to drift down irrelevant paths and corrupt the final inferred relation chain.

To this end, we propose \texttt{FlowRAG}, a semantic-aware retrieval framework designed to overcome these limitations through two key innovations. First, to address the entity sparsity issue, we construct a quad-level heterogeneous graph encompassing passages, summaries, sentences, and entities. By introducing summary nodes as coarse semantic hubs, \texttt{FlowRAG} bridges the gap between abstract queries and specific entities, ensuring robust activation even when explicit keywords are missing. Second, to mitigate noise and error propagation, we introduce a frequency-aware weighted flow mechanism. This module routes relevance through entity-passage links weighted by within-passage term frequency, effectively pruning noisy connections and extracting high-confidence reasoning paths as an explicit logic skeleton for generation. Extensive experiments show that our model outperforms existing strong baselines effectively.

Our contributions are summarized as follows:
\begin{itemize}[leftmargin=*, align=left]
    \item We propose a new framework termed \texttt{FlowRAG}, which constructs a quad-level heterogeneous graph, enabling a Dual-Granularity Activation mechanism that synergizes fine-grained sentence matching with coarse-grained summary alignment to resolve semantic sparsity.
    \item We propose a Frequency-Aware Weighted Flow Algorithm to replace implicit topological scoring, utilizing Term Frequency to filter noise and extract interpretable, explicit reasoning paths for multi-hop queries.
    \item We conduct extensive experiments demonstrates that \texttt{FlowRAG} outperforms state-of-the-art baselines on complex reasoning benchmarks, validating the effectiveness of integrating semantic relevance with explicit information flow.
\end{itemize}

\begin{figure*}
\centering
\includegraphics[width=1\textwidth]{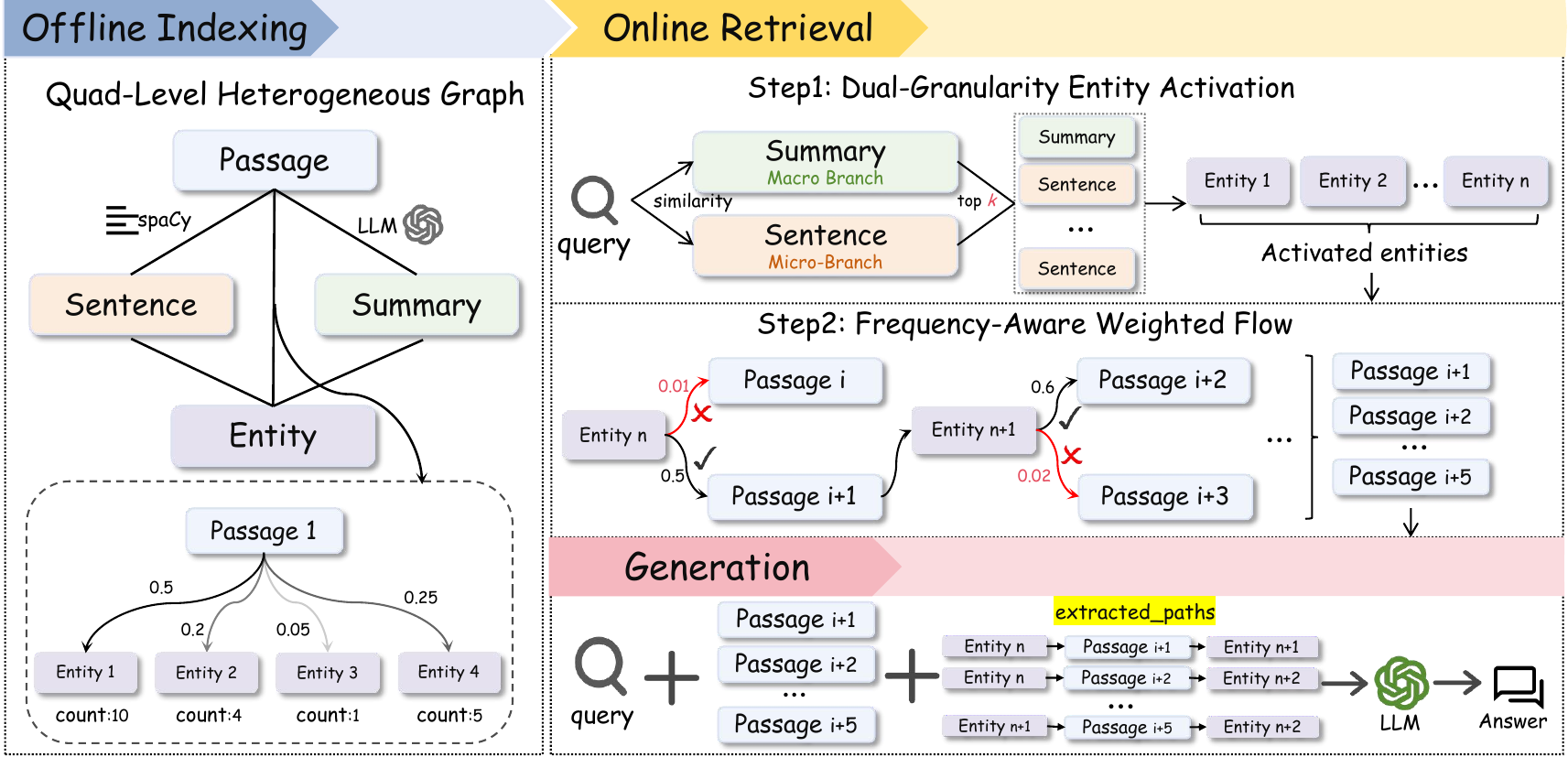} 
\caption{Overview of the \texttt{FlowRAG} framework. The framework consists of three main stages: 
(1) \textbf{Quad-Level Graph Construction}: We construct a heterogeneous graph incorporating Summary Nodes to bridge the semantic gap between high-level concepts and fine-grained details. 
(2) \textbf{Dual-Granularity Entity Activation}: The retrieval process initializes by matching the query against both Summary Nodes and Sentence Nodes to ensure comprehensive semantic coverage. 
(3) \textbf{Frequency-Aware Weighted Flow}: The frequency-aware weighted flow algorithm explicitly traces relevance along the edges to extract structured reasoning paths for the LLM.}
\label{fig:overview}
\vspace{-3mm}
\end{figure*}

\section{Related Work}

\subsection{Retrieval-Augmented Generation}
Retrieval-Augmented Generation (RAG) ~\cite{cao2025tv,lewis2020retrieval,cao2026agents} has emerged as a fundamental paradigm to mitigate hallucinations and knowledge obsolescence in Large Language Models (LLMs) by grounding generation in external corpora. Standard approaches predominantly rely on dense retrieval mechanisms, such as DPR ~\cite{karpukhin2020dense}, which map queries and passages into a shared latent space for semantic matching. To further enhance retrieval precision, subsequent research has introduced iterative retrieval strategies ~\cite{shao2023enhancing} to gather context over multiple steps, and self-reflective frameworks like CRAG ~\cite{yan2024corrective} and Self-RAG ~\cite{asai2024self}, which empower LLMs to actively critique, discard, or refine retrieved documents dynamically. 

Despite these advancements, these flat retrieval methods generally treat the corpus as a collection of independent text segments.  As a result, they inherently struggle to model explicit dependencies between disjointed documents, a limitation that hinders performance in complex multi-hop reasoning scenarios where bridging information gaps is essential.

\subsection{Graph Retrieval-Augmented Generation}
GraphRAG utilizes topological structures to capture entity dependencies essential for multi-hop reasoning. Early approaches focused on subgraph extraction or filtering via graph neural networks. For instance, KGP~\cite{wang2024knowledge} retrieves triples strictly relevant to query concepts, while G-retriever~\cite{he2024gretriever} and GFM-RAG~\cite{luo2025gfm} employ algorithms like Steiner trees to identify connected, semantically relevant subgraphs. Addressing multi-granular information, RAPTOR~\cite{sarthi2024raptor} introduces a recursive clustering mechanism to construct a hierarchical tree of summaries, capturing both thematic concepts and fine details through textual abstraction rather than explicit entity paths.

Recent methodologies have shifted towards mimicking associative memory for efficient indexing. HippoRAG~\cite{jimenez2024hipporag} and HippoRAG2~\cite{gutirrez2025ragmemorynonparametriccontinual} utilize Personalized PageRank to simulate neurobiological retrieval, discovering hidden connections without direct lexical overlap, while LightRAG~\cite{guo2024lightrag} enhances this with a dual-level indexing strategy for comprehensive coverage. Distinctly, LinearRAG~\cite{zhuang2025linearraglineargraphretrieval} proposes a streamlined, relation-free alternative using term co-occurrence to bypass expensive extraction pipelines. While efficient, its reliance on implicit scoring limits interpretability and precision in multi-hop constraints, highlighting the need for methods capable of extracting explicit reasoning flows.

\section{Methodology}
\subsection{Overview} 
In this paper, we propose \texttt{FlowRAG}, which improves the standard retrieval pipeline by moving from static document indexing to dynamic structural navigation. The framework has three coupled phases (Figure \ref{fig:overview}). First, in the offline indexing phase, we build a Quad-Level Heterogeneous Graph that explicitly encodes the corpus's semantic hierarchy. Unlike bipartite designs, this topology adds dense semantic representations to bridge the gap between raw text and abstract queries. Second, the online retrieval phase uses dual-granularity activation to start the search with relevance signals from both fine-grained textual details and coarse thematic summaries. Finally, to ensure interpretability and depth, we introduce a frequency-aware weighted flow algorithm to find explicit logical trajectories. It propagates query energy across the graph and filters noise with structural constraints, yielding coherent reasoning paths for the generator.

\subsection{Quad-Level Heterogeneous Graph Construction}
To fully capture the hierarchical semantics of the corpus, \texttt{FlowRAG} proposes a quad-level heterogeneous graph, denoted as $G = (V, E)$. This structure explicitly models the semantic hierarchy from macro-level summaries to micro-level entities. Specifically, the vertex set $V$ consists of four distinct layers of semantic granularity: 
\begin{equation}
    V = V_P \cup V_{Sum} \cup V_S \cup V_E,
\end{equation}
where $V_P$ represents the original passages; $V_{Sum}$ denotes dense semantic representations (summaries) generated for each passage by a Large Language Model; $V_S$ represents fine-grained sentences segmented from passages; and $V_E$ represents named entities extracted via lightweight Named Entity Recognition (NER) from both sentences and summaries. The edge set $E$ is constructed to capture semantic flow through four specific interaction types, formalized as adjacency matrices. 

First, to preserve the connection between abstract concepts and detailed content, we establish a binding matrix $B \in \mathbb{R}^{|V_P| \times |V_{Sum}|}$, where a strong bidirectional link is assigned ($B_{lk} = \lambda$) between a passage $p_k$ and its corresponding summary $sum_l$. Second, to capture macro-level themes, we define an Abstraction Matrix $A \in \{0,1\}^{|V_{Sum}| \times |V_E|}$, where $A_{lj} = 1$ if summary $sum_l$ mentions entity $e_j$. Third, to capture micro-level details, we define a Mention Matrix $M \in \{0,1\}^{|V_S| \times |V_E|}$, where $M_{ij} = 1$ if sentence $s_i$ mentions entity $e_j$.Finally, unlike LinearRAG, which treats passage-entity relations as binary, we pre-compute semantic relevance during indexing to construct a weighted Contain Matrix $C \in \mathbb{R}^{|V_P| \times |V_E|}$. We define the weight $C_{kj}$ using the normalized Term Frequency (TF) of entity $e_j$ within passage $p_k$:
\begin{equation}
    C_{kj} = \frac{\text{count}(e_j, p_k)}{\sum_{e' \in p_k} \text{count}(e', p_k)}.
\end{equation}This design retains the linear scalability of graph construction while enriching the topological information. By storing $C$ and $B$ as weighted sparse matrices while keeping $M$ and $A$ binary, we ensure efficient memory usage for large-scale corpora.

\begin{table*}[t]
\centering
\small %
\setlength{\tabcolsep}{1pt} %
\begin{tabular}{lccccccccc}
\toprule
\multirow{2}{*}{\textbf{Method}} & \multicolumn{2}{c}{\textbf{HotpotQA}} & \multicolumn{2}{c}{\textbf{2WikiMultiHopQA}} & \multicolumn{2}{c}{\textbf{MuSiQue}} & \textbf{Medical} & \multicolumn{2}{c}{\textbf{Avg}} \\
\cmidrule(lr){2-3} \cmidrule(lr){4-5} \cmidrule(lr){6-7} \cmidrule(l){8-8} \cmidrule(l){9-10}
 & Contain-Acc. & GPT-Acc. & Contain-Acc. & GPT-Acc. & Contain-Acc. & GPT-Acc. & GPT-Acc. & Contain-Acc. & GPT-Acc. \\
\midrule
Vanilla RAG & 55.70 & 58.60 & 48.60 & 43.00 & 26.10 & 29.60 & 61.68 & 43.47 & 48.22 \\
KGP & 61.50 & 60.90 & 31.60 & 30.00 & 25.60 & 30.10 & 54.22 & 39.57 & 43.81 \\
G-retriever & 42.20 & 40.60 & 46.60 & 27.10 & 14.40 & 15.50 & 50.36 & 34.40 & 33.39 \\
RAPTOR & 55.90 & 58.30 & 50.10 & 42.10 & 23.30 & 27.40 & 55.75  & 43.10 & 45.89\\
E$^2$GraphRAG & 61.00 & 63.90 & 54.30 & 38.10 & 23.80 & 26.20 & 58.00 & 46.37 &  46.55\\
LightRAG & 60.30 & 59.50 & 55.20 & 39.00 & 27.40 & 28.60 & 54.36 & 47.63 & 45.37\\
HippoRAG & 57.00 & 59.30 & 66.10 & 59.90 & 29.30 & 24.10 & 55.04 & 50.80 & 49.59 \\
GFM-RAG & 62.70 & 65.60 & 66.80 & 59.60 & 29.90 & 34.60 & 56.07 & 53.13 & 53.97\\
HippoRAG2 & 62.90 & 64.30 & 62.70 & 55.00 & 31.00 & 35.00 & 60.77 & 52.20 & 53.77\\
LinearRAG & \underline{65.50} & \underline{66.80} & \underline{68.90} & \underline{62.70} & \textbf{32.30} & \underline{35.10} & \underline{64.06} & \underline{55.57} & \underline{57.17}\\
\midrule
\textbf{\texttt{FlowRAG} (ours)} & \textbf{66.30} & \textbf{68.60} & \textbf{71.20} & \textbf{65.20} & \underline{32.20} & \textbf{37.10} & \textbf{64.65} & \textbf{56.57} & \textbf{58.89} \\
\bottomrule
\end{tabular}
\vspace{-1mm}
\caption{Result (\%) of baselines and \texttt{FlowRAG} on four benchmark datasets in terms of Contain-Match and GPT-Evaluation Accuracy. The best and second results for each dataset are highlighted with \textbf{bold} and \underline{underline}.}
\label{tab1}
\vspace{-3mm}
\end{table*}

\subsection{Dual-Granularity Entity Activation}
Direct entity matching often misses relevant contexts when queries are abstract or utilize different terminology. To address this, we propose a Dual-Granularity Activation mechanism that synthesizes relevance signals from both micro-level details and macro-level themes. Given a query $q$ with embedding $\mathbf{h}_q$, we compute an activation vector $\mathbf{a}^{(0)} \in \mathbb{R}^{|V_E|}$. For each candidate entity $e_j$, the score is derived from the maximum activation of two parallel branches:
\begin{equation}
    a_j^{(0)} = \max \left( \mathcal{S}_{micro}(q, e_j),\quad \mathcal{S}_{macro}(q, e_j) \right).
\end{equation}

The Micro-Branch ($\mathcal{S}_{micro}$) captures fine-grained keyword matching by utilizing the mention matrix $\mathbf{M}$ to propagate relevance from the top-$K$ most similar sentences, where $K$ is a hyperparameter controlling the activation breadth:
\begin{multline}
    \mathcal{S}_{micro}(q, e_j) = \\
    \max_{i: M_{ij}=1} \left( \cos(\mathbf{h}_q, \mathbf{h}_{s_i}) \cdot  \mathbb{I}(s_i \in \text{Top-}K) \right).
\end{multline}

The Macro-Branch ($\mathcal{S}_{macro}$) captures thematic alignment. It utilizes the Abstraction Matrix $\mathbf{A}$ to propagate relevance from the top-$K$ most similar summaries:
\begin{multline}
    \mathcal{S}_{macro}(q, e_j) =\\
    \max_{l: A_{lj}=1} \left( \cos(\mathbf{h}_q, \mathbf{h}_{sum_l}) \cdot \mathbb{I}(sum_l \in \text{Top-}K) \right),
\end{multline}
where $\mathbb{I}(\cdot)$ denotes the indicator function. This hybrid initialization ensures that the search is anchored not only by explicit mentions but also by implicit thematic connections.

\subsection{Frequency-Aware Weighted Flow}
We move from activated entities to passages and model retrieval as explicit energy propagation, not an implicit random walk. Inspired by PathRAG~\cite{chen2025pathragpruninggraphbasedretrieval}, we propose a Frequency-Aware Weighted Flow algorithm that uses precomputed semantic relevance to guide reasoning.

The transition weight $W_{u \to v}$ measures the semantic conductivity of an edge. For entity-to-passage transitions, we use the TF-weighted Contain Matrix $\mathbf{C}$ to favor passages where the entity is a core topic rather than an incidental mention:
\begin{equation}
    W_{e_j \to p_k} = \frac{C_{kj}}{\sum_{p' \in \mathcal{N}(e_j)} C_{p'j}}.
\end{equation}

The flow propagates iteratively. Let $R_t(u)$ be the residual energy at node $u$ at hop $t$. Propagation to a neighbor $v$ at hop $t+1$ follows the recursive equation:
\begin{equation}
    R_{t+1}(v) = R_t(u) \cdot \alpha \cdot W_{u \to v} \cdot \mathbb{I}(R_{t+1}(v) > \tau),
\end{equation}
where $\alpha \in (0,1)$ denotes a decay factor that controls depth and limits drift on long paths. $\tau$ denotes a dynamic pruning threshold that removes low-confidence paths.

Finally, \texttt{FlowRAG} extracts explicit reasoning paths $P = (e_{start} \to \dots \to p_{target})$. In this way, we score a path by the average energy flux along its sequence as follows:
\begin{equation}
    \text{Score}(P) = \frac{1}{|P|} \sum_{v \in P} R(v).
\end{equation}

We select the top paths to form a logic skeleton, giving the large language model structured, verified evidence to reduce hallucination.

\section{Experiments}
\subsection{Experimental Setting}
\paragraph{Datasets.} To rigorously evaluate the effectiveness of \texttt{FlowRAG}, we conduct experiments on four comprehensive benchmarks, including three widely adopted multi-hop QA datasets: HotpotQA~\cite{yang2018hotpotqa}, 2WikiMultiHopQA~\cite{ho2020constructing}, and MuSiQue~\cite{trivedi2022musique}, as well as the domain-specific Medical dataset from GraphRAG-Bench~\cite{xiang2025use}. To ensure a fair comparison with state-of-the-art baselines, we strictly align our experimental protocol with HippoRAG and LinearRAG. Specifically, we utilize the standard retrieval corpus associated with each dataset and evaluate performance on a consistent sample of 1,000 questions drawn from the respective validation sets.

\paragraph{Baselines.}
We compare \texttt{FlowRAG} against a diverse array of retrieval methods. First, we evaluate Vanilla RAG utilizing dense retrieval with CoT prompting. Second, we benchmark against leading GraphRAG Systems, specifically KGP~\cite{wang2024knowledge}, G-retriever~\cite{he2024gretriever}, RAPTOR~\cite{sarthi2024raptor}, E$^2$GraphRAG~\cite{zhao2025e2graphragstreamlininggraphbasedrag}, LightRAG~\cite{guo2024lightrag}, HippoRAG~\cite{jimenez2024hipporag}, GFM-RAG~\cite{luo2025gfm}, and HippoRAG2~\cite{gutirrez2025ragmemorynonparametriccontinual}. Finally, we explicitly utilize LinearRAG as a distinct baseline. Since LinearRAG shares the relation-free philosophy but relies on implicit PageRank, this comparison is critical for demonstrating the effectiveness of our proposed weighted flow algorithm.

\paragraph{Evaluation Metrics.}
We evaluate our method \texttt{FlowRAG} using four metrics across two categories, following the evaluation of LinearRAG~\cite{zhuang2025linearraglineargraphretrieval}. For end-to-end QA performance, we use: 1) Contain-Match Accuracy (Contain-Acc.), which checks if the correct answer appears in the generated response, and 2) GPT-Evaluation Accuracy (GPT-ACC.), an LLM-based metric that assesses whether the predicted answer matches the ground truth. Note that for the Medical dataset, we only evaluate using GPT-ACC due to the lengthy nature of golden answers. Regarding retrieval quality, we incorporate metrics from GraphRAG-Bench~\cite{xiang2025use}: Context Relevance, measuring the semantic alignment between questions and retrieved passages, and Evidence Recall, evaluating whether the retrieved contents contain all necessary information.

\paragraph{Implementation Setup.} For consistency in implementation, we adopt all-mpnet-base-v2~\cite{song2020mpnet} as the underlying embedding model for all algorithms. The number of retrieved documents is fixed at $k=5$ for all retrieval-based methods. Additionally, GPT-4o-mini\footnote{https://platform.openai.com/docs/models/gpt-4o-mini} serves as the uniform large language model for both response generation and performance evaluation.

\subsection{Main Results}
Table \ref{tab1} presents the comparative performance of \texttt{FlowRAG} against state-of-the-art baselines. Overall, \texttt{FlowRAG} establishes a new state-of-the-art, achieving the highest Avg scores in both metrics and surpassing the strongest baseline, LinearRAG, by 1.72\% in GPT-Acc. This demonstrates the superiority of integrating dense semantic representations with explicit information flow.

Specifically, compared to structure-aware methods (e.g., LightRAG, HippoRAG), \texttt{FlowRAG} shows robust improvements, particularly on multi-hop reasoning tasks. For instance, on 2WikiMultiHopQA, \texttt{FlowRAG} surpasses HippoRAG by a large margin of 5.3\% in GPT-Acc. Furthermore, against the strong relation-free baseline LinearRAG, \texttt{FlowRAG} exhibits superior reasoning capabilities, achieving absolute GPT-Acc gains of 2.1\% on HotpotQA and 2.5\% on 2WikiMultiHopQA. This advantage stems from our architectural duality: while LinearRAG relies on micro-level sentence matching, \texttt{FlowRAG}'s dual-granularity activation captures high-level thematic signals often missed by sentence-only retrievers. Crucially, our explicit flow mechanism filters irrelevant edges, resolving the semantic sparsity issue that limits the reasoning depth of implicit scoring methods.

\begin{figure}[t!]
\centering
\includegraphics[width=0.48\textwidth]{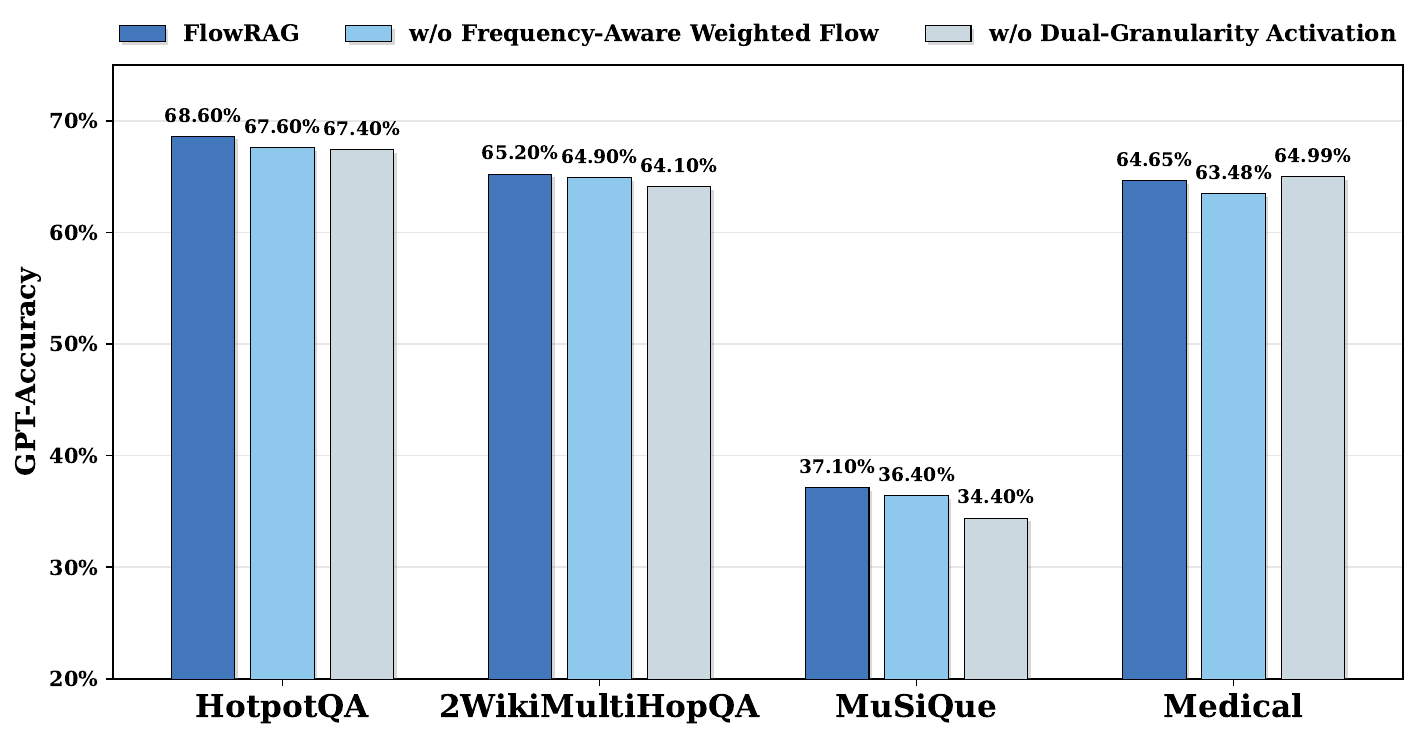} 
\vspace{-5mm}
\caption{Ablation study on key modules of \texttt{FlowRAG} under four different datasets.}
\label{fig:ablation}
\vspace{-4mm}
\end{figure}

Notably, the results on the MuSiQue dataset highlight the superior noise-robustness of our approach. While LinearRAG achieves a marginally higher Contain-Acc (32.30\% vs. 32.20\%), \texttt{FlowRAG} significantly outperforms it in the final answer generation (GPT-Acc: 37.10\% vs. 35.10\%). This discrepancy indicates that LinearRAG optimizes for high recall at the cost of introducing substantial noise into the context window. In contrast, \texttt{FlowRAG} prioritizes information density via weighted flow paths. This results in a cleaner context that, despite a trivial drop in span recall, significantly enhances the LLM's ability to extract correct answers. Additionally, the peak performance on the Medical dataset (64.65\%) confirms that our weighted construction generalizes effectively to specialized domains without relying on complex, pre-defined ontologies.

\subsection{Ablation Studies}
To verify the contribution of each core component, we conducted an ablation study across four datasets, as illustrated in Figure \ref{fig:ablation}. The removal of Dual-Granularity Activation results in a substantial performance drop on multi-hop reasoning benchmarks, with GPT-Accuracy decreasing by 2.7\% on MuSiQue and 1.1\% on 2WikiMultiHopQA. This decline confirms that relying solely on micro-level sentence matching exacerbates semantic sparsity, whereas our macro-level summary nodes effectively bridge the granularity gap for abstract queries. Interestingly, on the Medical dataset, removing this module yields a marginal improvement (+0.34\%), suggesting that in highly specialized domains requiring precise terminological alignment, coarse-grained summaries may introduce slight abstraction noise, making fine-grained entity matching more effective.

Conversely, the Frequency-Aware Weighted Flow mechanism proves critical for robustness across all domains. Eliminating this module leads to consistent degradation, including drops of 1.0\% on HotpotQA and 1.17\% on the Medical dataset. This underscores the necessity of explicit, density-guided propagation; without the filtering provided by Term Frequency weights, the retrieval process becomes vulnerable to irrelevant connections and noise. Consequently, the full \texttt{FlowRAG} framework achieves the optimal balance, demonstrating that explicit structure-aware flow is essential for extracting high-quality reasoning paths while dual-granularity activation is key for handling complex semantic dependencies.

\begin{figure}
\centering
\includegraphics[width=0.48\textwidth]{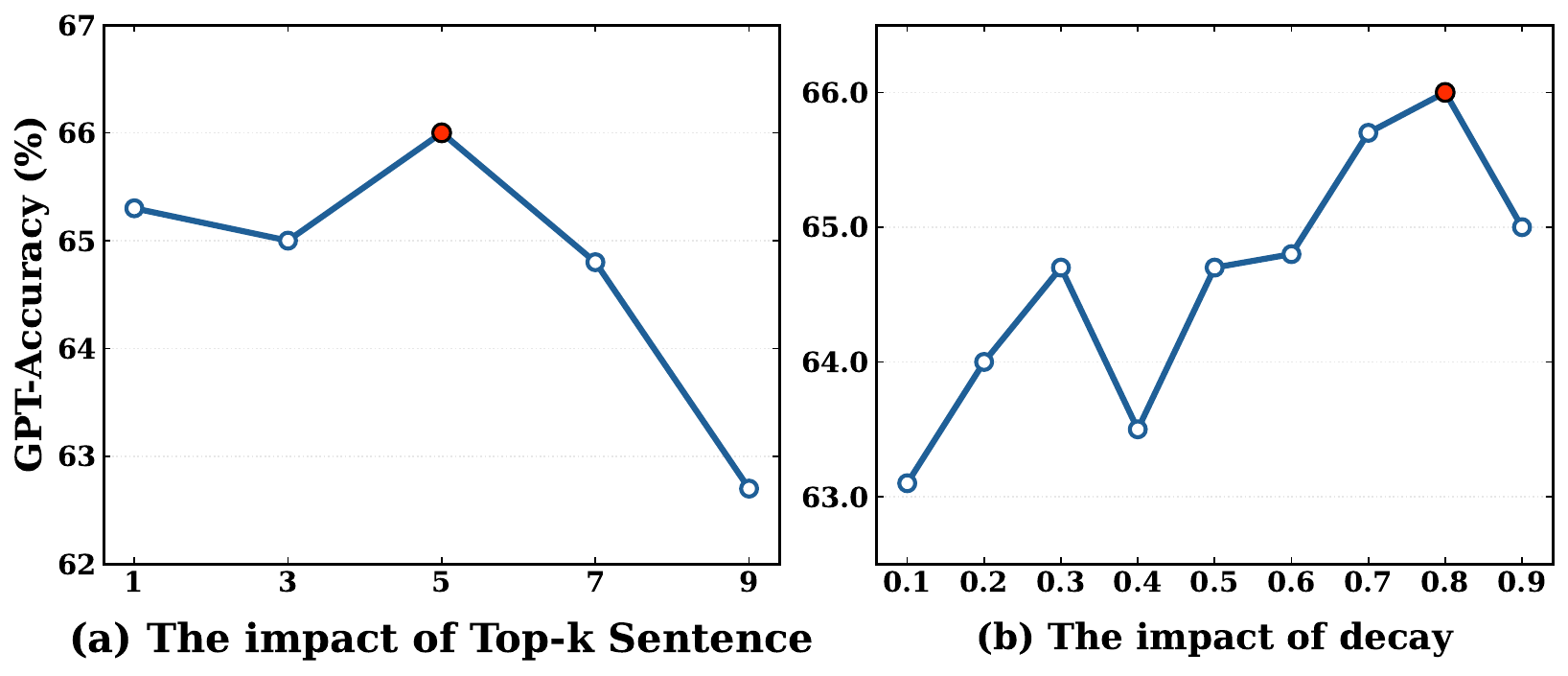} 
\vspace{-5mm}
\caption{Hyper-Parameter analysis of \texttt{FlowRAG} performance in the 2WikiMultiHopQA dataset.}
\label{fig:analysis}
\vspace{-4mm}
\end{figure}

\begin{table*}[htbp]
\centering
\small 
\setlength{\tabcolsep}{3pt} 
\begin{tabular}{lcccccccccc}
\toprule
\multirow{2}{*}{\textbf{Method}} & \multicolumn{2}{c}{\textbf{Fact Retrieval}} & \multicolumn{2}{c}{\textbf{Complex Reasoning}} & \multicolumn{2}{c}{\textbf{Contextual}} & \multicolumn{2}{c}{\textbf{Creative Generation}}  & \multicolumn{2}{c}{\textbf{Avg}} \\ 
\cmidrule(lr){2-3} \cmidrule(lr){4-5} \cmidrule(lr){6-7} \cmidrule(lr){8-9} \cmidrule(lr){10-11}
& Recall & Relevance & Recall & Relevance & Recall & Relevance & Recall & Relevance & Recall & Relevance\\ 
\midrule
Vanilla RAG         & 86.24 & 63.71 & 84.97 & \textbf{84.11} & 84.14 & \textbf{89.94} & 44.88 & \underline{58.73} & 75.06   & \underline{74.12}\\ 
RAPTOR              & 85.40 & 69.38 & 89.70 & 53.20 & 88.86 & 58.73 & 72.70 & 52.71 & 84.16    & 58.50 \\ 
E$^2$GraphRAG       & 87.84 & 69.74 & \underline{87.08} & 62.67 & \underline{89.17} & 71.63 & 60.26 & 35.84 & 81.09   &     59.97\\ 
LightRAG            & 80.32 & 41.27 & 82.91 & 42.79 & 85.71 & 43.11 & 81.34 & 45.17 & 82.57  &  43.09\\ 
HippoRAG            & 87.25 & 52.44 & 83.80 & 42.19 & 83.46 & 49.13 & 81.66 & 45.03 & 84.04    & 47.20\\ 
GFM-RAG             & \underline{90.08} & 57.90 & 85.03 & 33.06 & 78.62 & 40.14 & 83.51 & 22.87 & 84.31   &    38.49\\ 
LinearRAG           & 88.86 & \textbf{86.09} & 87.03 & \underline{81.58} & 89.13 & \underline{87.89} & \underline{89.08} & \textbf{72.74} & \underline{88.52}   &    \textbf{82.08}\\ 
\midrule
\textbf{\texttt{FlowRAG} (ours)}  & \textbf{90.25} & \underline{80.44} & \textbf{91.89} & 66.99 & \textbf{94.66} & 81.83 & \textbf{94.81} & 50.00 & \textbf{92.90}   &   69.82\\ 
\bottomrule
\end{tabular}
\vspace{-1mm}
\caption{Retrieval quality evaluation results (\%) of different baselines across four different question categories. The table compares recall and relevance metrics for fact retrieval, complex reasoning, contextual understanding, and creative generation tasks. The best results are highlighted in \textbf{bold}, and second-best results are \underline{underlined}.}
\label{tab:retrieval_results}
\vspace{-1mm}
\end{table*}

\begin{table*}[t]
\centering
\small 
\setlength{\tabcolsep}{4pt} 
\begin{tabular}{lccccccc}
\toprule
\multirow{2}{*}{\textbf{Model}} & \multicolumn{2}{c}{\textbf{HotpotQA}} & \multicolumn{2}{c}{\textbf{2WikiMultiHopQA}} & \multicolumn{2}{c}{\textbf{MuSiQue}} & \textbf{Medical} \\
\cmidrule(lr){2-3} \cmidrule(lr){4-5} \cmidrule(lr){6-7} \cmidrule(l){8-8}
 & Contain-Acc. & GPT-Acc. & Contain-Acc. & GPT-Acc. & Contain-Acc. & GPT-Acc. & GPT-Acc. \\
\midrule
all-mpnet-base-v2 & \textbf{66.30} & 68.60 & 71.20 & 65.20 & 32.20 & 37.10 & \textbf{64.65} \\
all-MiniLM-L6-v2  & 66.10 & 67.10 & \textbf{71.60} & \textbf{66.40} & 32.20 & 36.10 & 64.31 \\
bge-large-en-v1.5 & 66.20 & \textbf{69.30} & 70.70 & 65.70 & \textbf{34.60} & \textbf{38.00} & 62.90 \\
e5-large-v2       & 63.20 & 64.30 & 69.90 & 64.70 & 30.50 & 35.20 & 61.35 \\
\bottomrule
\end{tabular}
\vspace{-1mm}
\caption{Comparison of different sentence embedding models used in \texttt{FlowRAG}.}
\label{tab:embedding_results}
\vspace{-3mm}
\end{table*}

\subsection{Hyper-Parameter Analysis}
We analyze the sensitivity of our model to two key hyperparameters: the number of top-$k$ sentences and the decay factor. Specifically, Figure \ref{fig:analysis} illustrates the impact of these parameters on the 2WikiMultiHopQA dataset in terms of GPT-Accuracy.

\textbf{Impact of Top-$k$ Sentence.} 
As shown in Figure \ref{fig:analysis}(a), we investigate how the number of retrieved sentences affects model performance. The value of $k$ is varied from 1 to 9 with a step size of 2. The accuracy initially increases, reaching a peak of $66.0\%$ at $k=5$. This indicates that sufficient context is crucial for the model's reasoning. However, when $k$ exceeds 5, the performance begins to degrade significantly, dropping to $62.7\%$ at $k=9$. This decline suggests that introducing an excessive number of sentences may include irrelevant noise, which hinders the model's prediction capability.

\textbf{Impact of Decay Factor.} 
Figure \ref{fig:analysis}(b) demonstrates the effect of the decay parameter, ranging from 0.1 to 0.9 with a step of 0.1. The results show a fluctuating but generally upward trend as the decay value increases. The model performance starts at $63.1\%$ (decay=0.1) and experiences a temporary drop at 0.4 ($63.5\%$). The optimal performance is achieved at a decay value of 0.8, yielding the highest accuracy of $66.0\%$. Beyond this optimal point, the accuracy decreases slightly to $65.0\%$ when the decay is set to 0.9. Based on these observations, we set the decay factor to 0.8 for our experiments.

\subsection{Retrieval Quality Analysis}
Table \ref{tab:retrieval_results} presents the comparative results of retrieval quality. \texttt{FlowRAG} exhibits a significant advantage in information coverage, achieving the highest Recall across all four categories with an average of 92.90\%. Notably, in the challenging \texttt{Contextual} and \texttt{Creative Generation} tasks, \texttt{FlowRAG} outperforms the strong baseline \texttt{LinearRAG} by margins of 5.53\% and 5.73\% in Recall, respectively. This confirms that our flow-based mechanism effectively traverses complex graph structures to uncover implicit connections that dense retrieval methods often miss.
While \texttt{LinearRAG} achieves the highest Relevance scores (Avg 82.08\%), indicating a high density of useful information per retrieved chunk, \texttt{FlowRAG} adopts a ``high-coverage" strategy (Avg Recall 92.90\%). While LinearRAG achieves the highest Relevance scores (Avg 82.08\%), indicating a high density of useful information per retrieved chunk, \texttt{FlowRAG} adopts a "high-coverage" strategy (Avg Recall 92.90\%). 

\textcolor{black}{We argue that this apparent lower relevance is a deliberate and beneficial consequence of our retrieval objective. In multi-hop scenarios, optimizing strictly for per-chunk retrieval relevance often fragments the reasoning chain by missing critical bridging facts. \texttt{FlowRAG} intentionally prioritizes "reasoning relevance" by retrieving a broader context to ensure complete evidence coverage. While this introduces slightly less relevant chunks, modern LLMs possess strong capabilities to filter this noise, whereas they cannot recover from broken evidence chains caused by low recall. Furthermore, when evaluating the recall-relevance frontier by treating the number of retrieved chunks ($k$) as a budget variable, \texttt{FlowRAG} proves highly efficient. At the same budget of $k=5$, \texttt{FlowRAG} captures significantly more critical evidence (92.90\% Recall) than the strongest baselines, demonstrating a superior allocation of the retrieval budget for complex reasoning.}

\begin{table*}[t]
\centering
\begin{tabular}{lcccccc}
\toprule
\textbf{} & \textbf{Indexing} & \textbf{Retrieval} & \textbf{Path Ext.} & \textbf{Prompt} & \textbf{Compl.} & \textbf{Acc.} \\
 & \textbf{Time (s)} & \textbf{Time (s)} & \textbf{Time (s)} & \textbf{Tokens ($10^6$)} & \textbf{Tokens ($10^6$)} & \textbf{(\%)} \\ \midrule
HippoRAG & 936.00 & 1.461 & -- & 3.05 & 0.98 & 63.00 \\
HippoRAG2 & 1147.01 & 1.694 & -- & 4.98 & 1.22 & 58.85 \\
LightRAG & 4933.22 & 10.963 & -- & 35.52 & 51.16 & 47.10 \\ \midrule
\textbf{FlowRAG (Ours)} & \textbf{347.09} & \textbf{0.250} & \textbf{1.205} & \textbf{0.75} & \textbf{0.03} & \textbf{65.20} \\ \bottomrule
\end{tabular}
\caption{\textcolor{black}{Efficiency and cost comparison on 2Wiki.}}
\label{tab:efficiency}
\end{table*}

\subsection{Effectiveness of Different Sentence Embeddings}
We evaluate the impact of the embedding model on \texttt{FlowRAG}'s performance by benchmarking four representative models: all-mpnet-base-v2~\cite{song2020mpnet}, all-MiniLM-L6-v2~\cite{wang2020minilm}, bge-large-en-v1.5~\cite{xiao2024c}, and e5-large-v2~\cite{wang2022text}. As presented in Table \ref{tab:embedding_results}, the empirical results underscore that the choice of embedding critically influences both retrieval effectiveness and downstream generation accuracy. Notably, bge-large-en-v1.5 exhibits robust performance in complex multi-hop reasoning tasks, securing the highest GPT-Accuracy on HotpotQA (69.30\%) and MuSiQue (38.00\%). Meanwhile, all-mpnet-base-v2 proves highly effective in specialized domains, achieving a top score of 64.65\% on the Medical dataset. Remarkably, the lightweight all-MiniLM-L6-v2 surpasses larger counterparts on the 2WikiMultiHopQA dataset, achieving superior performance in both containment and generation metrics. In contrast, e5-large-v2 generally trails behind across most benchmarks. These findings suggest that while \textit{bge} and \textit{mpnet} offer strong representations for challenging and domain-specific queries, respectively, optimal model selection remains sensitive to the inherent characteristics of each dataset.

\subsection{\textcolor{black}{Efficiency and Cost Analysis}}
\textcolor{black}{To quantify the computational and operational overhead, we evaluated the indexing time, retrieval latency, and token consumption of \texttt{FlowRAG} against leading graph-based baselines on the 2WikiMultiHopQA dataset. As shown in Table~\ref{tab:efficiency}, \texttt{FlowRAG} demonstrates significant efficiency in offline indexing, operating approximately 2.7 times faster than HippoRAG and nearly 14 times faster than LightRAG. During online retrieval, our sparse matrix implementation yields an average latency of 0.25s. Furthermore, \texttt{FlowRAG} is highly cost-effective; it consumes only about 25\% of the prompt tokens required by HippoRAG and roughly 2\% of those required by LightRAG, while maintaining higher accuracy.}

\subsection{\textcolor{black}{Robustness and LLM Scaling}}
\textcolor{black}{\texttt{FlowRAG} mitigates sensitivity to summary quality through structural robustness. First, our dual-granularity activation provides a failover mechanism: even if a summary omits details, the sentence-level micro-branch ensures critical entities are still activated. Second, the frequency-aware weighted flow acts as an implicit quality filter; if a summary introduces a hallucinated entity, its low term frequency in the original passage results in a near-zero edge weight that prunes the connection. Additionally, as shown in Table~\ref{tab:scaling}, \texttt{FlowRAG} scales positively with stronger LLMs. Using GPT-4o further boosts performance across all benchmarks, confirming its adaptability to frontier-level models.}

\begin{table}[t]
\small
\centering
\setlength{\tabcolsep}{0.8pt}
\begin{tabular}{lccc}
\toprule
\textbf{} & \textbf{HotpotQA} & \textbf{2Wiki} & \textbf{MuSiQue} \\
 & \textbf{(GPT-Acc)} & \textbf{(GPT-Acc)} & \textbf{(GPT-Acc)} \\ \midrule
\textbf{FlowRAG (GPT-4o)} & \textbf{71.60} & \textbf{66.50} & \textbf{39.00} \\ \bottomrule
\end{tabular}
\caption{\textcolor{black}{Performance scaling with GPT-4o.}}
\label{tab:scaling}
\end{table}

\subsection{Case Study: Disambiguating Complex Familial Relations}
\label{sec:case_study}



To rigorously evaluate the reasoning fidelity of \texttt{FlowRAG} in dense knowledge scenarios, we conduct a deep analysis of a representative multi-hop query from the 2WikiMultihopQA dataset (see Table \ref{tab:deep_case_study}). This case challenges the model to traverse a specific lineage path: identifying the father-in-law of ``Princess Elisabeth, Duchess of Hohenberg.''

The baseline, LinearRAG, incorrectly identifies ``Archduke Franz Ferdinand'' as the answer. This failure illustrates a phenomenon we term \textit{Semantic Distraction}. In the dense vector space, the embedding for ``Duke of Hohenberg'' is overwhelmingly dominated by the historically prominent figure Franz Ferdinand (the subject's grandfather-in-law), overshadowing the less famous correct entity, Maximilian. The vector retriever effectively performs a ``fuzzy match'' based on historical co-occurrence popularity rather than precise kinship constraints. Consequently, the model conflates the query subject with the most popular entity in the semantic neighborhood, ignoring the specific generational relationship required by ``father-in-law.''

\begin{table}[t!]
    \centering
    \small
    \renewcommand{\arraystretch}{1.5}
    \begin{tabular}{p{0.95\linewidth}}
        \hline
        \textbf{Question:} Who is Princess Elisabeth, Duchess Of Hohenberg's father-in-law? \\
        \textbf{Gold Answer:} Maximilian, Duke of Hohenberg \\
        \hline
        \textbf{LinearRAG} \\
        \textbf{Prediction:} textcolor{black}{Archduke Franz Ferdinand of Austria}\\
        \textbf{Retrieved Context:} The dense retrieval model returned passages with high semantic overlap for ``Hohenberg'' and ``Duke''. However, these passages were dominated by the most historically prominent figure, Archduke Franz Ferdinand (the subject's grandfather-in-law), overshadowing the less famous Maximilian. \\
        \textit{Failure Mode:} \textbf{Semantic Distraction \& Title Conflation}. The model succumbed to the strong co-occurrence of ``Hohenberg'' with the famous Archduke, ignoring the precise distinction between the titles ``Duke'' and ``Archduke,'' as well as the generational constraint. \\
        \hline
        \textbf{\texttt{FlowRAG} (Ours)} \\
        \textbf{Prediction:} \textcolor{blue}{Maximilian, Duke of Hohenberg}\\
        \textbf{Key Extracted Paths (from Graph):} \\
        1. \texttt{hohenberg} $\xrightarrow{0.324}$ \texttt{[Doc 215]} $\xrightarrow{rel}$ \texttt{sophie} \\
        2. \texttt{hohenberg} $\xrightarrow{0.323}$ \texttt{[Doc 215]} $\xrightarrow{rel}$ \texttt{austria} \\
        \textit{Success Mechanism:} \textbf{Structural Anchoring}. As shown in the extracted paths, \texttt{FlowRAG} identified the critical intermediate entity \texttt{Sophie} (the first Duchess of Hohenberg) via Document 215. By grounding the reasoning in this specific lineage node, the model successfully established the correct family branch, navigating one generation down to correctly identify her son, Maximilian. \\
        \hline
    \end{tabular}
    \caption{Deep analysis of case study.}
    \label{tab:deep_case_study}
\end{table}


In contrast, \texttt{FlowRAG} successfully retrieves ``Maximilian, Duke of Hohenberg.'' This success stems from the model's ability to \textit{explicitize latent structure} rather than relying solely on semantic similarity. As shown in the extracted paths (Table \ref{tab:deep_case_study}), \texttt{FlowRAG} prioritizes a specific topological path:
\begin{equation*}
    \text{Subject} \xrightarrow{\text{context}} \textbf{Sophie} \xrightarrow{\text{lineage}} \text{Maximilian}
\end{equation*}
By explicitly traversing through the intermediate node \textit{Sophie} (the first Duchess of Hohenberg), the model establishes a structural anchor. This node acts as a disambiguation filter, effectively grounding the reasoning in the correct morganatic branch of the family tree. This structural constraint prevents the hallucination of semantically similar but factually incorrect ``popular'' entities, demonstrating that \texttt{FlowRAG}'s graph-based paths serve as a logical scaffold that ensures reasoning fidelity even when semantic signals are noisy or biased.

\section{Conclusion}
We present \texttt{FlowRAG}, a structure-aware framework that bridges static ranking and dynamic reasoning. It combines a Quad-Level Heterogeneous Graph with a Frequency-Aware Weighted Flow Algorithm to resolve granularity mismatches and extract explicit reasoning chains. Extensive experiments show that \texttt{FlowRAG} outperforms state-of-the-art baselines on complex multi-hop benchmarks, showing that combining semantic relevance with weighted information flow is effective. It also improves downstream generation by providing interpretable logic skeletons that reduce hallucinations. In future work, we will extend this flow-based mechanism to dynamic knowledge updates and test it in broader open-domain settings.

\section*{Acknowledgements}
This work was done during his internship at Shanghai Artificial Intelligence Laboratory. This work is supported by Shanghai Artificial Intelligence Laboratory. This work is funded by the National Nature Science Foundation of China (No.62477010, No.62577022 and No.62307028), Shanghai Science and Technology Innovation Action Plan (No.24YF2710100), and CIPS-SMP-Zhipu Large Model Fund.

\section*{Limitations}
While \texttt{FlowRAG} achieves superior reasoning fidelity, we acknowledge minor trade-offs inherent to its structural depth. First, the construction of the Quad-Level Graph and the execution of the Weighted Flow Algorithm introduce a moderate computational overhead compared to simple flat indexing, primarily due to LLM-based summarization and graph propagation. However, this is a necessary cost for enabling explicit reasoning. Second, our frequency-based pruning strategy, while effective generally, may benefit from further refinement in highly specialized domains where rare terms are pivotal. Finally, the current implementation focuses on static knowledge bases; extending the framework to handle real-time dynamic updates represents a promising direction for future optimization.

\section*{Ethical considerations}
The three datasets used in the experiments, including HotpotQA, 2WikiMultiHopQA, and MuSiQue are widely used datasets. The medical benchmark dataset is built from public resources. Our research strictly adheres to the Code of Ethics, particularly regarding data privacy, transparency, and responsible computing practices. There are no participants involved.

\bibliography{custom}

\appendix


\end{document}